\documentclass[letterpaper]{article} 
\usepackage{aaai25}  
\usepackage{times}  
\usepackage{helvet}  
\usepackage{courier}  
\usepackage[hyphens]{url}  
\usepackage{graphicx} 
\usepackage{natbib}  
\usepackage{caption} 
\usepackage{algorithm}
\usepackage{algorithmic}
\usepackage{booktabs}
\usepackage{amsmath}
\usepackage{multirow}
\usepackage{amssymb}

\usepackage[utf8]{inputenc} 
\usepackage[T1]{fontenc}    
\usepackage{url}            
\usepackage{amsfonts}       
\usepackage{nicefrac}       
\usepackage{microtype}      
\usepackage{xcolor}         
\usepackage{graphicx}         %
\usepackage{xurl}
\usepackage{float}
\usepackage{lipsum}
\usepackage{colortbl}
\usepackage{makecell}
\usepackage{xcolor}
\usepackage{wrapfig}
\usepackage{newfloat}
\usepackage{listings}
\DeclareCaptionStyle{ruled}{labelfont=normalfont,labelsep=colon,strut=off} 
\floatstyle{ruled}
\newfloat{listing}{tb}{lst}{}
\floatname{listing}{Listing}
\title{Efficient Reinforcement Learning Through Adaptively Pretrained Visual Encoder}
\author{
    Yuhan Zhang\textsuperscript{\rm 1}\textsuperscript{\rm 2}\equalcontrib
    Guoqing Ma\textsuperscript{\rm 1}\textsuperscript{\rm 3}\equalcontrib,
    Guangfu Hao\textsuperscript{\rm 1}\textsuperscript{\rm 2},
    Liangxuan Guo\textsuperscript{\rm 1}\textsuperscript{\rm 3},
    Yang Chen\textsuperscript{\rm 1}\textsuperscript{\rm 4},
    Shan Yu\textsuperscript{\rm 1}\textsuperscript{\rm 3}\textsuperscript{\rm 4}\thanks{Corresponding author.}
}
\affiliations{
    \textsuperscript{\rm 1}Laboratory of Brain Atlas and Brain-inspired Intelligence, Institute of Automation, Chinese Academy of Sciences\\
    \textsuperscript{\rm 2}School of Artificial Intelligence, University of Chinese Academy of Sciences\\
    \textsuperscript{\rm 3}School of Future Technology, University of Chinese Academy of Sciences\\
    \textsuperscript{\rm 4}Key Laboratory of Brain Cognition and Brain-inspired Intelligence Technology, Chinese Academy of Sciences\\

    \{zhangyuhan2022, maguoqing2022, haoguangfu2021, guoliangxuan2021, yang.chen, shan.yu\}@ia.ac.cn
}

\usepackage{bibentry}

\begin{document}

\maketitle

\begin{abstract}
While Reinforcement Learning (RL) agents can successfully learn to handle complex tasks, effectively generalizing acquired skills to unfamiliar settings remains a challenge. One of the reasons behind this is the visual encoders used are task-dependent, preventing effective feature extraction in different settings. To address this issue, recent studies have tried to pretrain encoders with diverse visual inputs in order to improve their performance. However, they rely on existing pretrained encoders without further exploring the impact of pretraining period. In this work, we propose APE: efficient reinforcement learning through \textbf{A}daptively \textbf{P}retrained visual \textbf{E}ncoder—a framework that utilizes adaptive augmentation strategy during the pretraining phase and extracts generalizable features with only a few interactions within the task environments in the policy learning period. Experiments are conducted across various domains, including DeepMind Control Suite, Atari Games and Memory Maze benchmarks, to verify the effectiveness of our method. Results show that mainstream RL methods, such as DreamerV3 and DrQ-v2, achieve state-of-the-art performance when equipped with APE. In addition, APE significantly improves the sampling efficiency using only visual inputs during learning, approaching the efficiency of state-based method in several control tasks. These findings demonstrate the potential of adaptive pretraining of encoder in enhancing the generalization ability and efficiency of visual RL algorithms.
\end{abstract}

%

\section{Introduction}

Deep Reinforcement Learning (Deep RL) has made great advances in recent years. Notable algorithms such as MuZero \cite{MuZero}, Player of Games \cite{PlayerofGame} and ReBeL \cite{ReBel} have been proposed to solve many challenging decision making problems.
While these advances have primarily focused on state-based inputs, significant progress has also been made in visual RL, i.e., leveraging image inputs for policy learning \cite{CURL, Dreamer, DreamerV2, dreamerv3, Kostrikov2020ImageAI}. 

However, visual RL agents learning from these high-demensional observations suffer from problems of low efficiency and often overfitting to specific environments \cite{Song2019ObservationalOI, CURL}.
Since the performance of these agents depends heavily on the quality of extracted features, the critical role of enhancing visual encoders has been highlighted in both model-free and model-based algorithms \cite{Yarats2019ImprovingSE,Dreamer, Poudel2023ReCoReRC}. 
\begin{figure}[t]
\centering
\includegraphics[width=0.9\columnwidth]{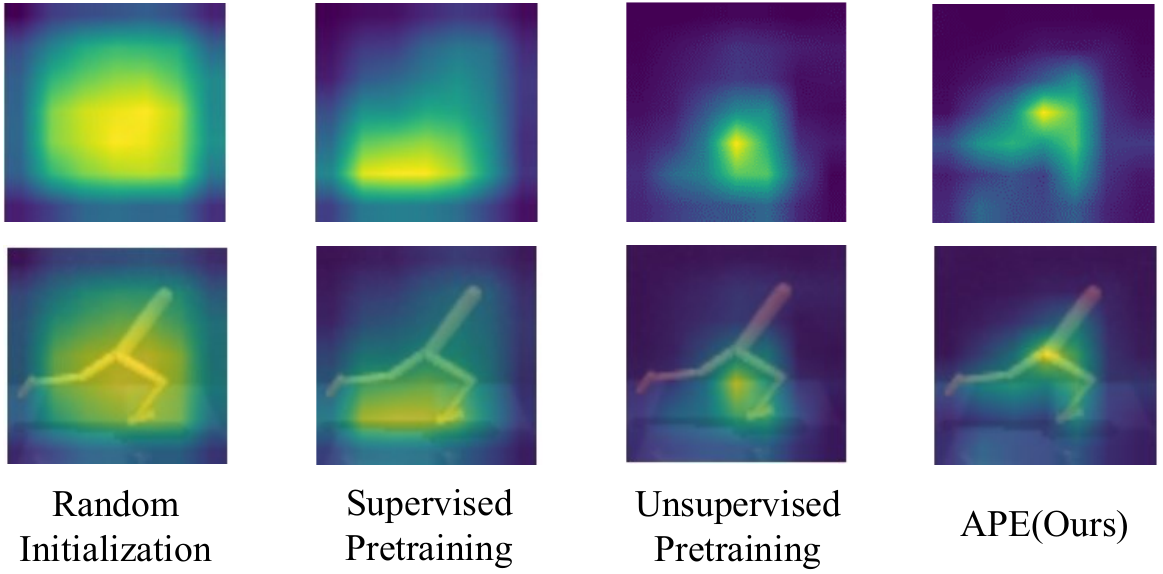} 
\caption{Visualization of ResNet-18 model with different pretraining strategy using LayerCAM \cite{9462463}, which indicates that APE is able to extract more precise outline of the Walker than other initialization settings. The first row displays the pure feature maps, which are also presented together with the image in the second row.}
\label{fig:vis init}
\end{figure}

In visual RL, various approaches have been explored to improve representation learning, among which data augmentations are often used to increase data diversity \cite{Wang2020ImprovingGI, Raileanu2021AutomaticDA,  10.1145/3510414, Liu2023RobustRL}. 
The challenge lies in extracting generalizable features rather than focusing on task-specific details, leading to difficulties in transferring learned skills to unseen scenarios \cite{Lee2019NetworkRA, Laskin2020ReinforcementLW}.

One promising direction is to exploit cross-domain knowledge learned by pretrained models \cite{RRL, Yuan2022PreTrainedIE}, which has shown great success in improving data efficiency and generalization ability in recent deep learning \cite{Devlin2019BERTPO, Baevski2020wav2vec2A}. In computer vision, since these models have typically been trained on extensive sets of natural images, their features inherently possess general knowledge about the world \cite{Hu2023ForPV}. This approach has the potential to enable RL agents to extract useful features more effectively, enhancing their ability to learn and generalize across different domains. Unsupervised learning, e.g., contrastive learning, is particularly advantageous in this regard as it enables pretrained models to extract meaningful features from unlabeled visual data, effectively addressing the issue of data scarcity and high labeling costs \cite{MoCo, SimCLR}.

Nevertheless, current RL methods simply implement existing pretrained models as visual encoders and augment observations in the downstream policy learning period \cite{RRL, Hu2023ForPV}.
As illustrated in Fig. \ref{fig:vis init}, the features learned by image classification models with the prevailing pretraining strategies (shown in the left three columns) exhibit limited generalization capabilities.
This also results in a lack of exploration of pretraining augmentations, which prove to be an important factor when applying pretrained encoders under great distribution shifts \cite{Geirhos2021PartialSI, Burns2023WhatMP}.

Given this, here we propose APE, a framework where the RL agent learns efficiently through \textbf{A}daptively \textbf{P}retrained visual \textbf{E}ncoder. This novel framework uses an adaptive closed-loop augmentation strategy in contrastive pretraining to learn transferable representations from a wide range of real-world images. Comparison in Fig. \ref{fig:vis init} indicates that APE helps to extract more generalizable features than other pretraining strategies. In addition, it works efficiently, requiring minimal interactions with the targeted environment during policy learning period. We evaluate our method on various challenging visual RL domains, including DeepMind Control (DMC) Suite \cite{dmc}, the Atari 100K benchmark \cite{Bellemare2012TheAL}, and Memory Maze \cite{MemoryMaze}. Experiments demonstrate that APE significantly improves the sampling efficiency and performance of the base RL method. Intersetingly, we found that the real RL enviorments are not necessary to test the pretrained encoder. Linear probes, a common protocol for evaluating the quality of learned representations \cite{SimCLR}, can serve as a useful metric to assess the quality of pretrained encoders quite effectively. 
The main contribution of this paper can be summarized as follows:

\begin{itemize}
    \item We propose a cross-domain RL framework with a fixed encoder pretrained on a wide variety of natural images using adaptive augmentation adjustment. This helps to produces more generalizable representations for the downstream RL tasks.
    
    \item We demonstrate the generality of APE to both model-based and model-free methods, underscoring its adaptability and effectiveness in enhancing learning performance across diverse RL approaches.
    
    \item APE is developed without any auxiliary tasks or other sensory informantion during policy learning period, effectively decoupling the pretraining phase from subsequent behavior learning tasks. This simple yet powerful design contributes to APE's superior performance on various visual RL benchmarks, approaching the performance of state-based Soft-Actor-Critic (SAC) \cite{SAC} in several control tasks.
\end{itemize}

\section{Related Works}

\subsection{Contrastive Learning}
In computer vision (CV), contrastive learning has gained popularity for its ability to learn generalizable representations leveraging unlabeled images and videos \cite{Oord2018RepresentationLW, SimCLR, MoCo}. Prior studies have emphasized the pivotal role of data augmentation in facilitating unsupervised training \cite{res7, res8, res9}.
Experiments conducted in SimCLR approach \cite{SimCLR} highlight the significant impact of data augmentations, which is re-confirmed by MoCo \cite{MoCo} and its modification MoCo v2 \cite{mocov2}.
AdDA \cite{Zhang2023AdaptiveDA} focuses on exploring the effect of dynamic adjustment on augmentation compositions, which enables the network to acquire more generalizable features. 
We adopt the feedback structure \cite{Zhang2023AdaptiveDA}  in the pretraining period and implement it on a different network architecture, which proves to be more suitable for RL tasks \cite{Yuan2022PreTrainedIE}.

\subsection{Representation Learning in RL}
There are extensive works in RL studying the impact of representation learning \cite{ Lin2020LearningTS, Liu2023RobustRL}, among which contrastive learning is often applied to acquire useful features \cite{Zhan2020LearningVR, Du2021CuriousRL, Schwarzer2021PretrainingRF}. CURL \cite{CURL} trains a visual representation encoder using contrastive loss, significantly improving sampling efficiency over prior pixel-based methods. Proto-RL \cite{Yarats2021ReinforcementLW} learns contrastive visual representations in dynamic RL environments without access to task-specific rewards. To make full use of context information, MLR \cite{Yu2022MaskbasedLR} introduces mask-based reconstruction to promote contrastive representation learning in RL. 
However, prior methods rely completely on data collected in target environments, which limits their generalization to unseen scenarios and hinders their adaptability to new tasks or environments. It also leads to additional sampling costs. APE, on the other hand, is pretrained on a distribution of real-world samples that wider than what policy can provide. 

Besides, the interpretability of extracted features is a key focus \cite{Lin2020SPACEUO, Delfosse2022BoostingOR, Delfosse2024InterpretableCB}, leading to improved performance and robustness of the agent. The efficiency gains of our method also result from a more interpretable encoder, aiding the agent in capturing key factors of observations in policy-making period.

\begin{figure*}[t]
\centering
\includegraphics[width=0.8\textwidth]{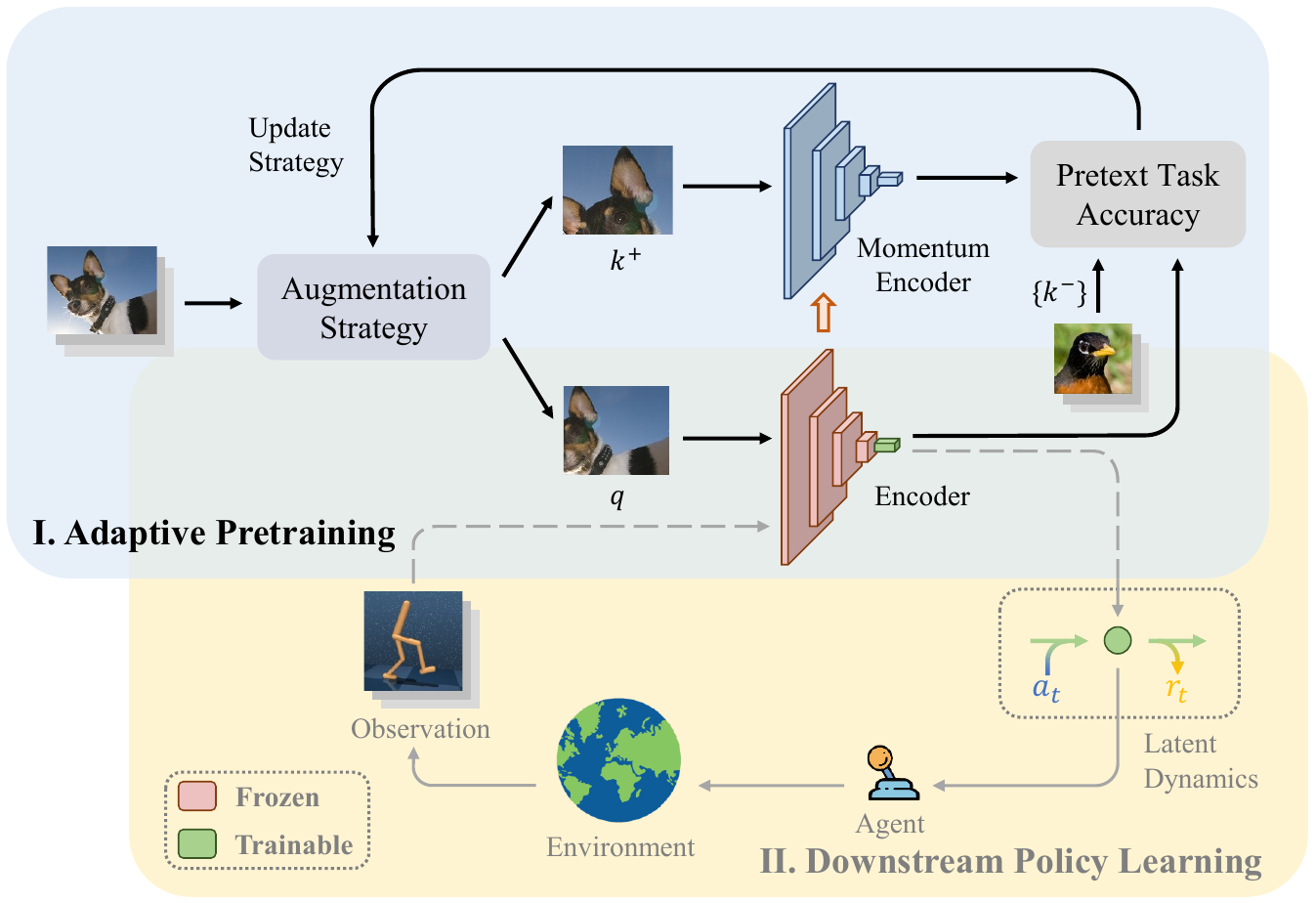} 
\caption{APE pipeline for MBRL. The training phase is divided into two parts, namely the Adaptive Pretraining period (within the 
blue area) and the Downstream Policy Learning period (within the 
yellow area). A wide variety of real-world images are augmented using an adaptive data augmentation strategy in the first period, which dynamically updates the sampling probability of each augmentation composition in the next pretraining epoch. In the second stage, the pretrained vision encoder is implemented in a generic RL framework as a perception module for the policy.}
\label{fig1}
\end{figure*}

\subsection{Generalization for Image-Based RL}
Since image augmentation has been successfully applied in CV for improving performance on object classification tasks, different approaches of transformation were investigated and incorporated in RL pipelines \cite{Laskin2020ReinforcementLW, Kostrikov2020ImageAI, Stooke2020DecouplingRL}. 
DrAC \cite{Raileanu2021AutomaticDA} contributes to the proper use of data augmentation for actor-critic algorithms and proposes an automatically selecting approach. 
SVEA \cite{Hansen2021StabilizingDQ} investigates the factors contributing to instability when employing augmentation within off-policy RL methods. DrQ \cite{Kostrikov2020ImageAI} together with DrQ-v2 \cite{Yarats2021MasteringVC} introduces a simple augmentation method for model-free RL algorithms utilizing input perturbations and regularization techniques, which we use to evaluate the generality of APE.
However, most previous methods attach more importance to the policy training period and straightforwardly augment the observations of the target environments \cite{Zhao2024AnEG}. Thus, they fall short in providing the requisite data diversity, which is essential for generalization over large domain gaps \cite{Yuan2022PreTrainedIE}. 
On the contrary, APE leverages an adaptively pretrained encoder without neglecting the potential benefits of pretraining augmentation strategy in RL, which has been confirmed in recent studies for its effectiveness in enhancing RL performance \cite{Burns2023WhatMP}.

\subsection{Pretrained Visual Encoders for RL}

Instead of training with expensive collected data, researches have also been made to bridge the domain gap between cross-domain datasets and the inputs of the target environments \cite{Ma2022VIPTU, Hu2023ForPV}. Using a pretrained ResNet encoder, RRL \cite{RRL} brings a straightforward approach to fuse extracted features into a standard RL pipeline. PIE-G \cite{Yuan2022PreTrainedIE} further demonstrates the effectiveness of supervised pretrained encoders by using early layer features from frozen models, with strongly augmented representations. By combining pretrained visual encoder and proprioceptive information, MVP outperforms supervised encoders in motor control tasks \cite{Xiao2022MaskedVP}. While pretrained models in aid of model-free RL have been studied, there lacks exploration on Model-Based Reinforcement Learning (MBRL) algorithms. These methods rely compeletely on reconstructed latents, thus further highlights the significance of representation learning \cite{Poudel2023ReCoReRC}.
Besides, extra tasks or sensory data are often needed during policy learning period while APE works without such intensive task-specific data.

\section{Preliminaries}
The proposed APE expands on both model-based and model-free RL methods. Detailed analyses are conducted on a mainstream MBRL framework, DreamerV3 \cite{dreamerv3}, which only learns from the representations extracted from original image observations. This integration allows APE to inherit DreamerV3's generality, operating with fixed hyperparameters across various domains. This section provides an overall description of our MBRL Backbone.
\subsubsection{Latent dynamics.}
The latent dynamics of DreamerV3 are modeled as a recurrent state space model (RSSM) which consists of the following five components:

\begin{figure*}[t]
\centering
\includegraphics[width=0.9\textwidth]{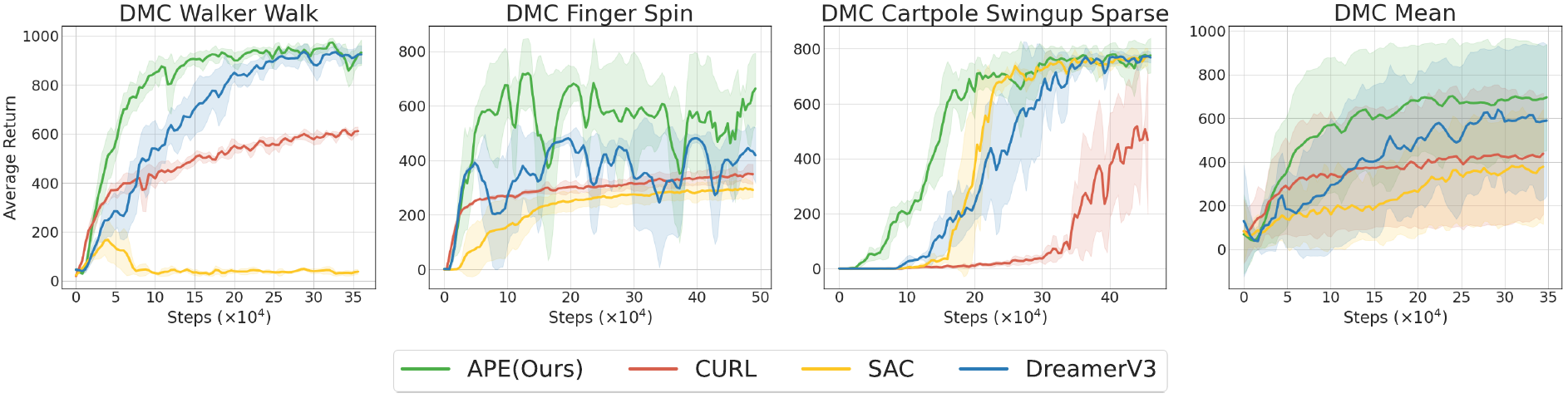} 
\caption{Training curves for DMC vision benchmarks.}
\label{figdmc}
\end{figure*}


\begin{equation}
\begin{aligned}
    \label{qua4}
    &\text{Encoder:} &z_t \sim f_{\theta}(z_t \mid z_{t-1},a_{t-1},o_t)\\
    &\text{Dynamics model:} &\hat{z}_t \sim p_{\theta}^{D}(\hat{z}_t \mid z_{t-1},a_{t-1})\\
    &\text{Reward predictor:} & \hat{r}_t \sim p_{\theta}^{R}(\hat{r}_t \mid z_t,z_{t-1},a_{t-1})\\
    &\text{Continue predictor:}& \hat{c}_t \sim p_{\theta}^{C}(\hat{c}_t \mid z_t,z_{t-1},a_{t-1})\\
    &\text{Decoder:}& \hat{x}_t\sim g_{\theta}(\hat{x}_t \mid z_t,z_{t-1},a_{t-1})
\end{aligned}
\tag{1}
\end{equation}

\noindent Here the dynamics model is designed to predict the next latent representation $\hat{z}_t$, while the feature ${z}_t$ generated by the encoder is used in the reward and continue predictor. The decoder using a convolutional neural network (CNN) helps in reconstructing visual inputs. 

\subsubsection{Agent learning.}
The actor-critic algorithm is employed to learn behaviors from the feature sequences predicted by the world model \cite{2018arXiv180310122H}. The actor aims to maximize the expected return $R_t$ for each state $s_t$ while the critic is trained to predict the return of each state $s_t$ with the current action $a_t$. Given $\gamma$ as the discount factor for the future rewards, the agent model are defined as follows:

\begin{equation}
\begin{aligned}
    \label{qua4}
    \text{Actor:} \quad & a_t \sim \pi_{\phi}(a_t \mid s_t)\\
    \text{Critic:} \quad &  V_{\psi} \approx  \mathbb{E}_{\pi_{\phi}, p_{\theta}}[\sum_{k=0}^{\infty}\gamma^kr_{t+k}]
\end{aligned}
\tag{2}
\end{equation}

The overall loss of the agent can be found in Appendix C.


\section{Methodology}

We consider the visual task as a Partially Observable Markov Decision Process (POMDP) \cite{Bellman1957AMD} due to the partial state observability from images. We denote the state space, the observation space, the action space and the reward function as $\mathcal{S, O, A }$ and $r$ respectively. The goal for an agent is to find a policy $\pi^*$ to maximize the expected cumulative return $E_p(\sum_{t=1}^Tr_t)$.
As shown in Fig. \ref{fig1}, our method decouples the pretraining period from the downstream control task and thus consists of two main parts: Adaptive Pretraining and Policy Learning, which are described as follows.

\subsection{Adaptive Pretraining}
Dynamic adjustment on data augmentation compositions is applied on MoCo v2 to explore the importance of visual encoder in RL methods. Instead of providing a complete search space for pretext task, APE provides the network with alternative compositions to learn robust and generalized representations. Specifically, two image features $q$ and $k^+$ extracted from two augmented views of a same image serve as a query \cite{MoCo} and a key. The set $\{k^-\}$ is made up of the outputs from other images as negative samples. For each augmentation composition, InfoNCE \cite{Oord2018RepresentationLW} is applied to maximize the agreement between $q$ and $k^+$:
\begin{equation}
     \ell_{q}=-\log{\frac{{\rm exp}(q\cdot k^+/\tau)}{{\rm exp}(q\cdot k^+/\tau)+\sum_{k^-}{\rm exp}(q\cdot k^-/\tau)} } 
\tag{3} 
\end{equation}

\noindent where $\tau$ is a temperature parameter and all the embeddings are $\ell_2$ normalized.
In our augmentation strategy, each batch is divided into $N$ sub-batches with the sampling probability $p_i$, i.e., $\sum_{i=1}^{N} p_i=1$, which is initialized as $1/N$ for a fair assignment.
The overall loss $\mathcal{L}_z$ of all the augmentation compositions is formulated as follows:

\begin{figure*}[t]
\centering
\includegraphics[width=0.9\textwidth]{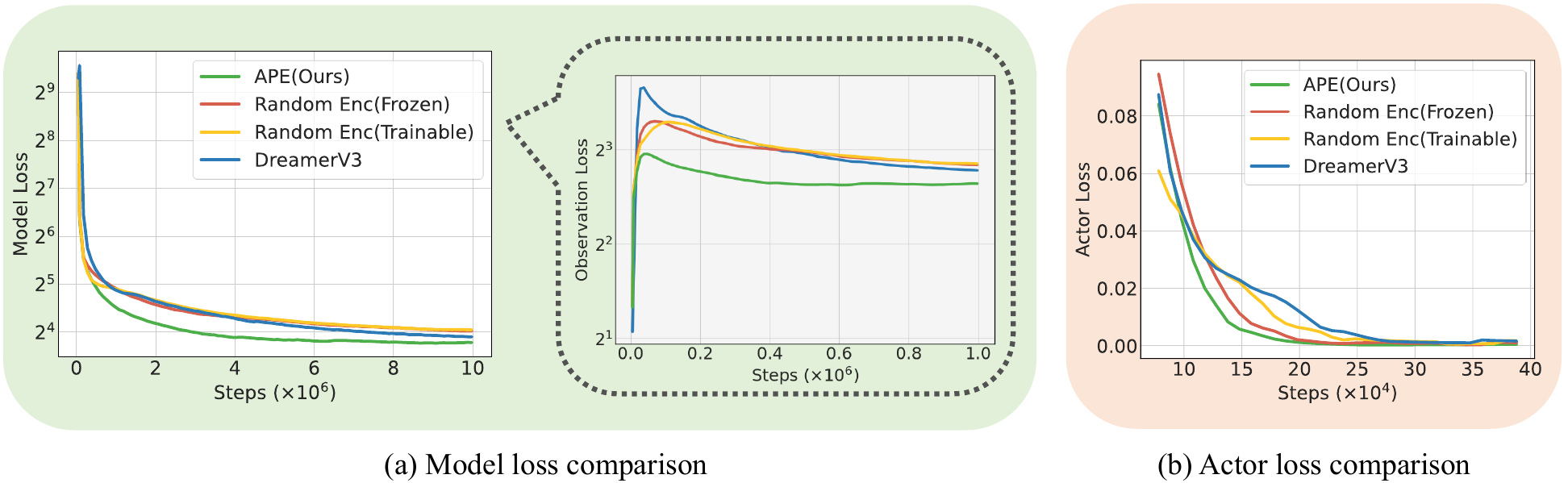} 
\caption{Loss comparison between DreamerV3, encoder with frozen random initialized parameters, encoder with trainable random initialized parameters and APE. The last layer of the frozen random initialized encoder is finetuned during training. The absolute value of actor loss is used.}
\label{figloss}
\end{figure*}

\begin{equation}
    \mathcal{L}_z=
    \begin{matrix} 
    \sum_{i=1}^{n} \ell_{q}p_i
    \end{matrix} 
    \tag{4} 
\end{equation}
\noindent Here $\mathcal{L}_z$ enables the encoder networks to maintain consistency across all sub-batches by utilizing the same key and query encoder.
The closed-loop feedback structure works by utilizing the sampling probability, which is dynamically updated at the end of every epoch by:
\begin{equation}
    \label{qua3}
    p^{t+1}=Softmax(\alpha(1-Acc^t))\tag{5} 
\end{equation}

\noindent where $\alpha$ is set to 0.8 for 7 compositions, and 1 for 3 compositions, thus speeds up the process of exploration when given more augmentation choices. This updating strategy decreases the size of those well-explored compositions and attaches more importance to the ones with lower pretext task accuracy in the next epoch.
\subsection{Policy Learning}

The pretrained encoder projects the high-dimensional image observations $o_t$ into low-dimensional latent features $z_t$, which are then transferred to the downstream agents that learn a control policy. The first three layers of the encoder are frozen to maintain generalization ability while parameters in the last layer are optimized together with the world model to adapt to environments with distribution shifts.

\label{wm}

\begin{figure}[t]
\centering
\includegraphics[width=0.9\columnwidth]{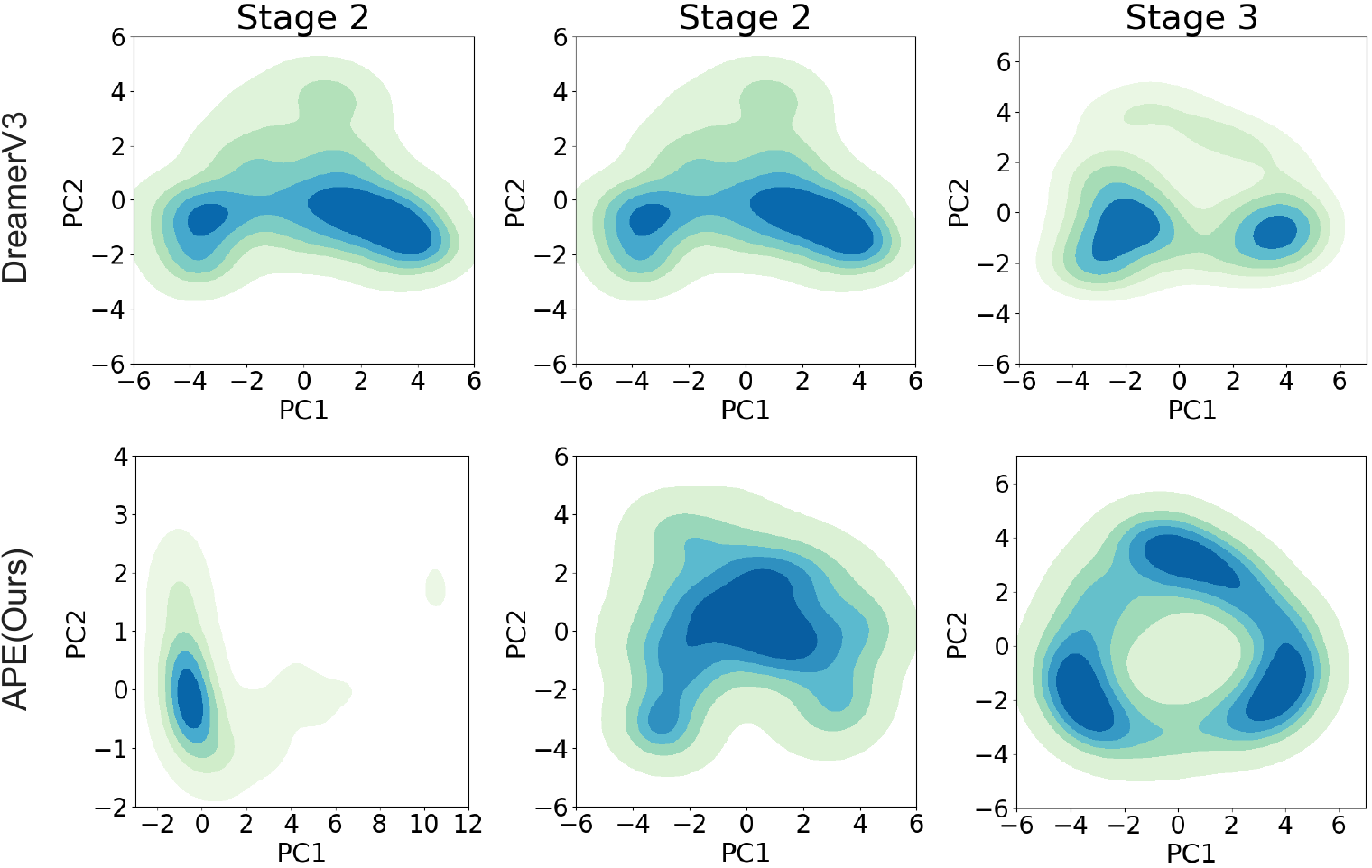} 
\caption{Exploration of states space in different phases during policy learning period. Data for 100 environment steps are sampled and visualized by Principal Component Analysis (PCA) in each stage. To compare fairly, axes are set to have identical ranges within the same stage. Thus the larger the state area, the higher the efficiency in exploration.}
\label{figstate}
\end{figure}

All model parameters $\theta$ in the latent dynamics except for the frozen ones in visual encoder's first three layers are optimized end-to-end to minimize the following objectives:

\begin{equation}
\setlength{\abovedisplayskip}{3pt}
\begin{aligned}
    \label{qualoss}
    \mathcal{L}_{rew}(\theta) = &-\log(p_{\theta}^{R}(\hat{r}_t \mid z_t,z_{t-1},a_{t-1}))\\
     \mathcal{L}_{con}(\theta)= &-\log(p_{\theta}^{C}(\hat{c}_t \mid z_t,z_{t-1},a_{t-1}))\\
     \mathcal{L}_{rec}(\theta)= &-\log(g_{\theta}(\hat{x}_t \mid z_t,z_{t-1},a_{t-1}))\\
     \mathcal{L}_{obs}(\theta) = & \beta_1 \max (1, \text{KL}[\text{sg}(f_{\theta}(z_t \mid z_{t-1},a_{t-1},o_t))\\
      & \parallel p_{\theta}^{D}(\hat{z}_t \mid z_{t-1},a_{t-1})])\\
     & + \beta_2 \max (1, \text{KL}[(f_{\theta}(z_t \mid z_{t-1},a_{t-1},o_t))\\
     & \parallel \text{sg}(p_{\theta}^{D}(\hat{z}_t \mid z_{t-1},a_{t-1}))])\\
\end{aligned}
\tag{6}
\end{equation}

\noindent where $\text{sg}(\cdot)$ denotes the stop-gradient operator. The fixed hyperparameters are set to $\beta_1 = 0.5$ and $\beta_2 = 0.1$.  
The overall loss of world model can be formulated as follows:

\begin{equation}
\setlength{\abovedisplayskip}{3pt}
    \label{qua4}
    \mathcal{L}(\theta)=\mathcal{L}_{rew}(\theta) + \mathcal{L}_{con}(\theta) + \mathcal{L}_{rec}(\theta) + \mathcal{L}_{obs}(\theta)
\tag{7}
\end{equation}


\begin{table}[h!]

  \begin{center}

    \label{table:pre} 

    \begin{tabular}{c|c|c} 
      \toprule  
      {Method}& $f_{\rm main}$ & {Acc. $(\%) $}\\
      \midrule
      \midrule
      MoCo v2 & — & 90.84\\
      APE &Jitter  & 91.08\\
       APE & Blur  & \textbf{91.7}\\

      \bottomrule 
    \end{tabular}
  \end{center}

  \caption{Comparison of different augmentation settings using linear probes on ImageNet-100 validation set. We report top-5 classification accuracy and bold the highest result. 
    }
    \label{table:linear}
\end{table}

Taking a multi-task view, the optimization of latent dynamics can be mainly divided into two parts, namely observation modeling and reward modeling \cite{HarmonyDreamTH}. APE works by contributing to the first modeling task, which is attached more importance in MBRL frameworks.

\begin{figure*}[t]
\centering
\includegraphics[width=0.9\textwidth]{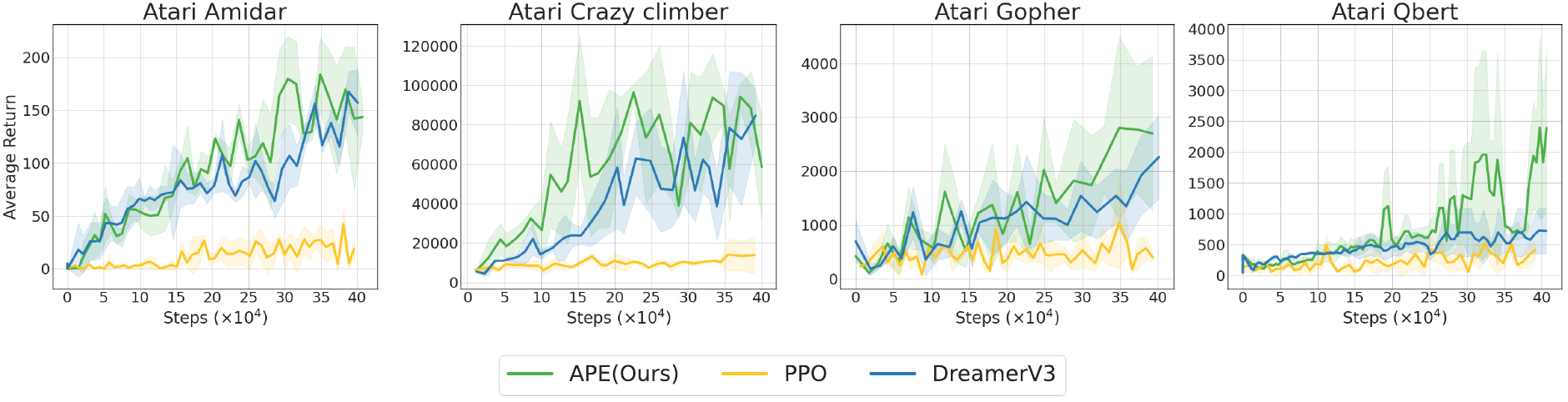} 
\caption{Training curves for Atari 100k benchmarks.}
\label{fig:Atari}
\end{figure*}

\begin{figure}[t]
\centering
\includegraphics[width=0.9\columnwidth]{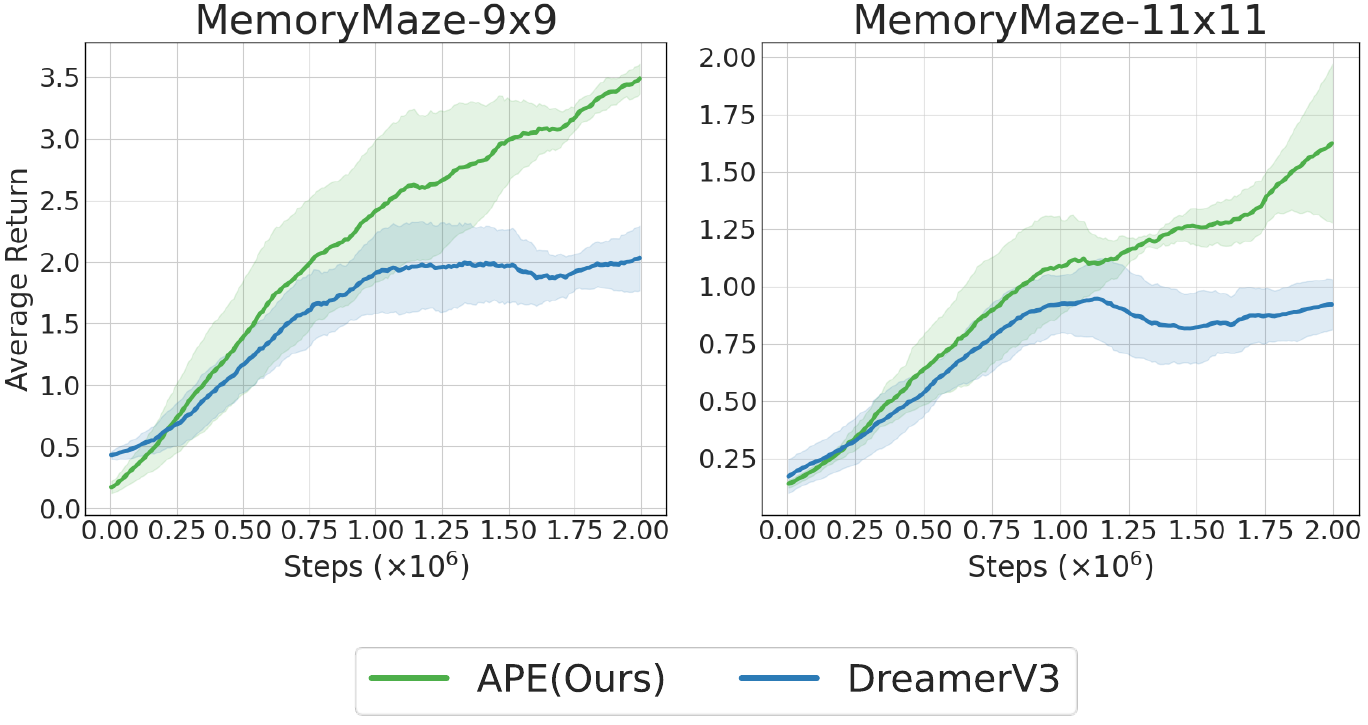} 
\caption{Training curves for Memory Maze benchmarks. }
\label{fig:mem}
\end{figure}

\begin{figure}[t]
\centering
\includegraphics[width=0.9\columnwidth]{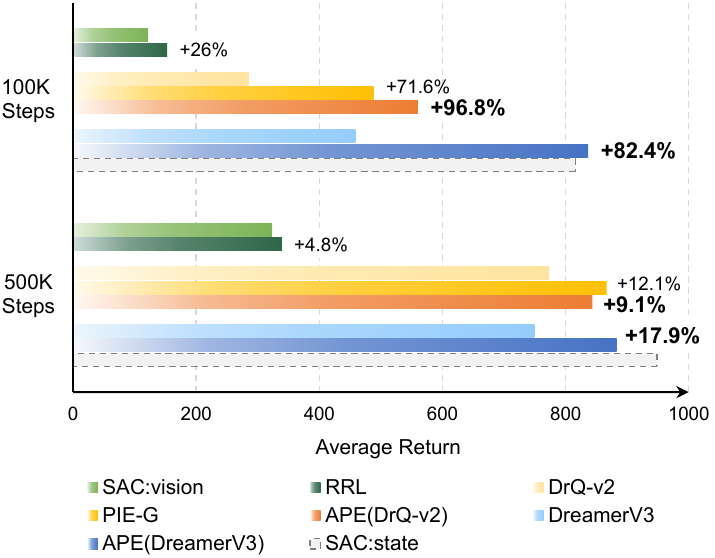} 
\caption{Comparison of DreamerV3-based and DrQ-v2-based APE against other ResNet pretrained algorithms (RRL and PIE-G), together with SAC:state, which learns on proprioceptive observations. 
The bars sharing the same color family (green, orange, and blue) denote algorithm groups following the same downstream strategy.
The performance gains are calculated based on the RL backbone of each group (SAC:vision, DrQ-v2 and DreamerV3), with APE showing the most significant improvement.
}
\label{figpieg}
\end{figure}

\section{Experiments}
Several experiments are conducted to evaluate the performance of APE using fixed hyperparameters, with details provided in Appendix A and C. We investigate the following questions: (a) Can APE improve the agent’s generalization ability and sampling efficiency on various visual RL benchmarks? (b) Can APE generalize to both model-
based and model-free methods? (c) Why APE works and how do choice of different settings affects the performance? By default, the encoder uses the ResNet18 architecture \cite{He2015DeepRL}. Results reported are averaged over at least 3 runs.

\subsection{Pretraining Encoders}
We pretrain APE on ImageNet-100, a randomly selected subset of the common ImageNet-1k \cite{5206848}, which has also been utilized in pervious works \cite{res21, Zhang2023AdaptiveDA} for pretext tasks.
Dynamic adjustment is made on the applied frequency of five data augmentations, including random color jittering, random grayscale conversion, random gaussian blur, random resized crop and random horizontal flip. Results under linear classification protocol are reported in Table \ref{table:linear}. The augmentation with varying applied frequency during pretraining is denoted as the main augmentation strategy ($f_{\rm main}$). In our method, the default $f_{\rm main}$ is random gaussian blur, which proved to be the most promising setting in AdDA.



\subsection{DMC Results}
Being a widely studied benchmark in control tasks, DMC provides a reasonably challenging and diverse set of environments. 
We evaluate the sample efficiency of APE on DMC vision tasks for 1M environment steps. As shown in Fig.\ref{figdmc}, experiments conducted on those tasks demonstrates that APE benefits from the strong feature extraction capabilities learned from ImageNet, leading to enhanced training efficiency and asymptotic performance when applied to control tasks.
Detailed comparison results of DMC scores are reported in Appendix B.

We illustrate the corresponding loss curves of \texttt{DMC walker walk} task learned with different encoders in Fig. \ref{figloss}. Encoders with random initialization have the same
network architecture as APE, with frozen or trainable random initialized parameters. Intuitively, a pretrained encoder helps accelerate the convergence of observation loss (shown in Fig. \ref{figloss}(a)), since it provides prior knowledge for extracting visual features. Moreover, model loss demonstrated in Fig. \ref{figloss}(a) indicates that APE also helps in the muti-task optimization of latent dynamics, as the overall model loss with pretrained encoder converges more easily than others. Besides, actor loss in Fig. \ref{figloss}(b) suggests that world model equipped with improved encoder is able to predict better future outcomes of potential actions, and thus speed up the actor's learning process.
Furthermore, by visualizing the states space in Fig. \ref{figstate}, we demonstrate that APE enables more sufficient exploration in states with larger visualization area, thereby enhancing the downstream performance. Visualization of reconstructions is provided in  Appendix B.

\subsection{Results on Other Benchmarks}

Fig. \ref{fig:Atari} indicates that APE achieves better or comparable performance using same hyperparameters on 4 Atari tasks. This environment is often used as a benchmark for investigating data-efficiency in RL algorithms. Following the common setup of Atari 100k, we set the environment steps to 40k in tasks considered. The performance on Atari benchmarks highlights the robustness and generalization capability of APE in various RL settings.

Additional experiments have also been made on Memory Maze, which is a 3D domain of randomized mazes generated from a first-person perspective, which measures the long-term memory of the agent and requires it to localize itself by integrating information over time. In this paper, tasks on Memory Maze are trained for 2M steps due to limited computational resources.
As shown in Fig. \ref{fig:mem}, APE is superior over the DreamerV3 baseline on these tasks that require semantic understanding of the environment, making it a promising candidate for complex real-world applications requiring sophisticated decision-making processes.

\subsection{Comparison with Other Pretrained Algorithms}
As shown in Fig. \ref{figpieg}, we compare the performances of two ResNet pretrained algorithms (RRL and PIE-G) and their base algorithms (SAC:vision and DrQ-v2) on three DMC benchmarks. APE outperforms all those methods on the 100K and 500K environment step benchmarks and achieves comparable performance with SAC:state (an agent that learns directly from states) at 100K environment step. To compare fairly, we reimplement DrQ-v2-based APE (denoted as APE (DrQ-v2)) to show that our findings and approach are not limited to MBRL framework.
The results of SAC:state, RRL and DrQ-v2 are from the paper of RRL \cite{RRL} and DrQ-v2 \cite{Yarats2021MasteringVC}, while the others are reproduced and averaged over at least 3 runs.
Detailed results are reported in the Appendix B.
\subsection{Ablation Studies}

\subsubsection{Pretraining does work.}
\begin{figure}[t]
\centering
\includegraphics[width=0.9\columnwidth]{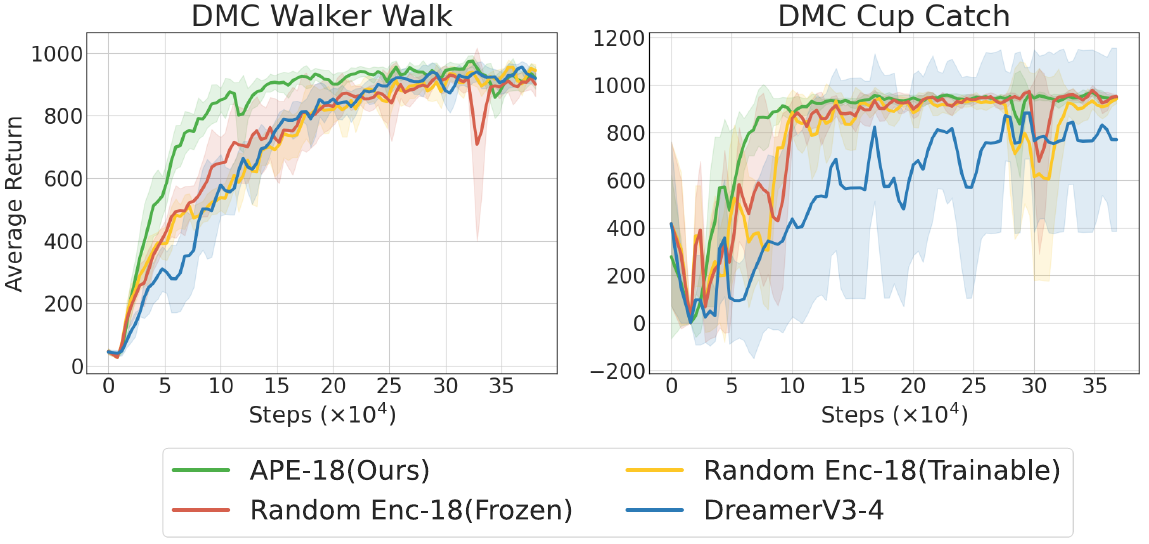} 
\caption{Choosing suitable pretraining strategy weighs more than increasing the depth of encoder network. We compare APE with random initialized encoder with frozen parameters, random initialized encoder with trainable parameters and DreamerV3. The last layer of the frozen random initialized encoder is finetuned during training. ‘-18’ and ‘-4’ denote the number of layers used in the encoder.}
\label{fig:init random}
\end{figure}

Experiments are conducted to figure out whether deeper encoder helps to extract more discriminative features (shown in Fig. \ref{fig:init random}). \textit{Random Enc}s with frozen or trainable initialized parameters have the same network architecture as APE and are included as baselines to eliminate the effect of varied network size. By comparing the performance of \textit{Random Enc} and DreamerV3, it is important to note that deeper networks do not always guarantee the extraction of better features, which leads to improved performance of APE in our tasks. This underscores the significant role of the pretraining period for RL algorithms.

\subsubsection{Augmentations matter.}

\begin{figure}[t]
\centering
\includegraphics[width=0.9\columnwidth]{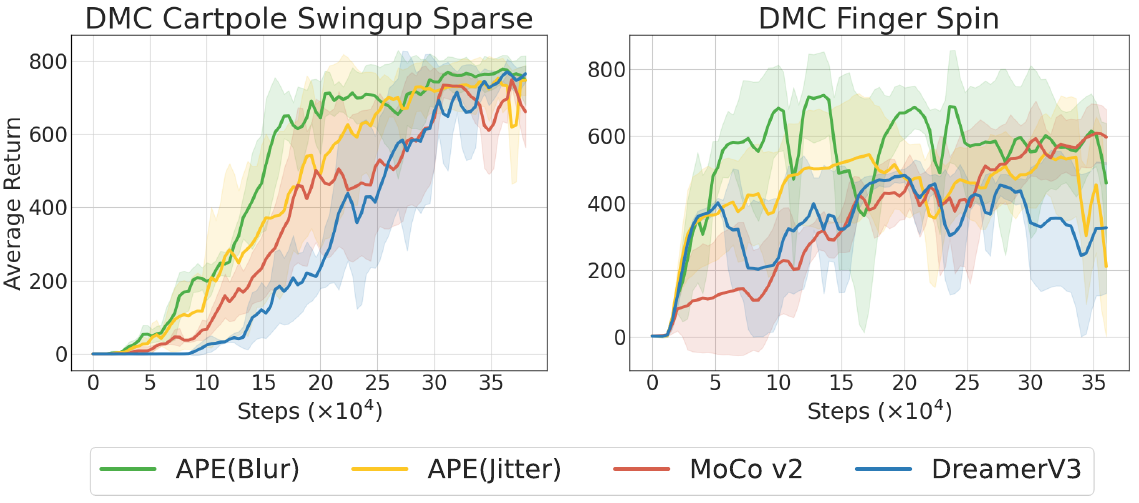} 
\caption{Different choices of augmentation strategy. APE with random gaussian blur as its main augmentation strategy outperforms other settings.}
\label{fig:aug}

\end{figure}

In Fig. \ref{fig:aug}, we focus on the applied frequency of random gaussian blur and random color jittering to investigate the effect of data agumentations on visual representations in RL tasks. We observe that the sampling efficiency varies when changing the augmentation strategy.
Results also indicates that linear probes may serve as a useful metric of pretrained model quality under the same network architecture, with relative findings made on imitation learning \cite{Hu2023ForPV}.

\begin{figure}[t]
\centering
\includegraphics[width=0.9\columnwidth]{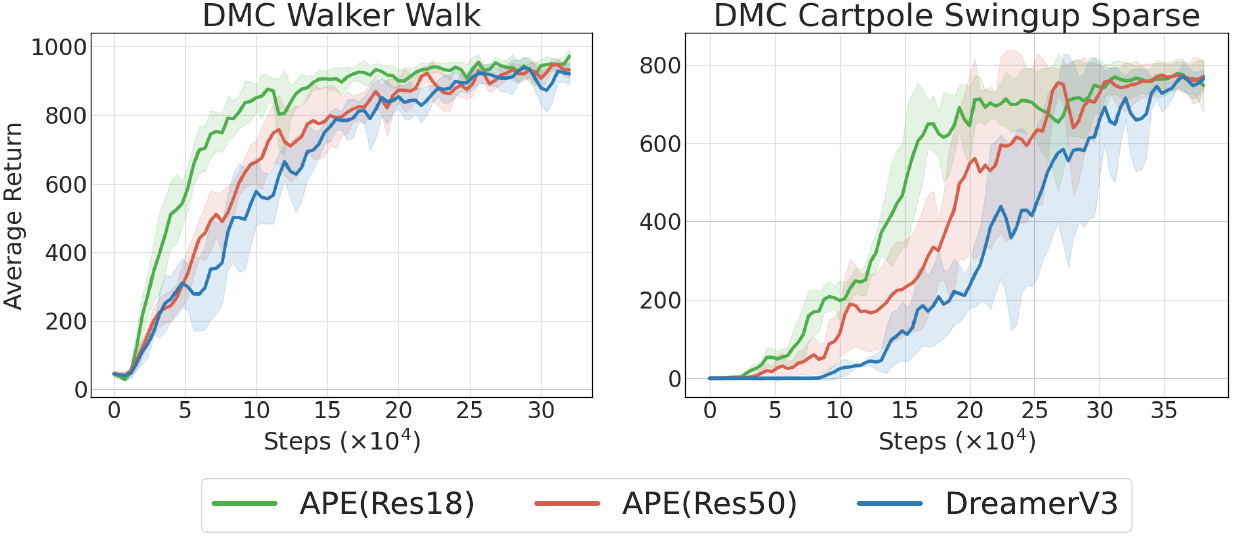} 
\caption{Different choices of network architectures. This figure indicates that APE with ResNet18 achieves better results compared with a deeper APE (ResNet50).}
\label{fig:arc}
\end{figure}

\subsubsection{Different choices of architectures.}

As shown in Fig. \ref{fig:arc}, we further explore the impact of architectures in DMC tasks. Following the settings of ResNet18 architecture, we freeze the first three layers of ResNet50 and update the last layer during training.
For a fair comparison, the latent dimension of the three architectures are kept same (4096) and both the ResNet18 and the ResNet50 architecture are pretrained on ImageNet-100 with the same $f_{\rm main}$.
Results demonstrate that the increase of depth and complexity of the network, which lead to more abstract representations, may compromises the performance of the fine-grained control tasks. 
Comparisons with ViT-based pretrained encoder \cite{He2021MaskedAA} are reported in the Appendix B.

\section{Conclusion}

In this paper, we propose APE, a simple yet effective method that implements adaptively pretrained encoder in RL frameworks. Unlike previous methods, APE is pretrained on a wide range of existing real-world images using a dynamic augmentation strategy, which helps the network to acquire more generalizable features in the downstream policy learning period.  
Experimental results show that our method surpasses state-of-the-art visual RL algorithms in learning efficiency and performance across various challenging domains. Besides, APE approaches the performance of state-based SAC in several control tasks, underscoring the effectiveness of augmentation strategy in the pretraining period.

\section{Acknowledgments}
This work was supported in part by the Strategic Priority Research Program of the Chinese Academy of Sciences (CAS)(XDB1010302), CAS Project for Young Scientists in Basic Research, Grant No. YSBR-041 and the International Partnership Program of the Chinese Academy of Sciences (CAS) (173211KYSB20200021).

\bigskip

\bibliography{aaai25}
\clearpage
\onecolumn
\appendix
\setcounter{figure}{0}
\setcounter{table}{0}
{\Large\textbf{Appendix of APE: Efficient Reinforcement Learning through Adaptively Pretrained Visual Encoder}}

\section{A\quad Environment}
\subsection{DeepMind Control (DMC) Suite \cite{dmc}}
Being a widely used RL benchmark, DMC contains a variety of continuous control tasks with a standardised structure and interpretable rewards. In this paper, we test the effectiveness of our method using DMC vision tasks, where the agent is required to learn low-level locomotion and manipulation skills operating purely from pixels. Visualized observations are in the first line of Fig. \ref{fig:env}.

\subsection{Memory Maze \cite{MemoryMaze}}
Agents in this benchmark is repeatedly tasked to navigate through randomized 3D mazes with various objects to reach. To succeed efficiently, agents must remember object locations, maze layouts, and their own positions. An ideal agent with long-term memory can explore each maze once and quickly find the shortest path to requested targets. 
The visualizations of the environment are shown in the second line of Fig. \ref{fig:env}, with \texttt{Agent Inputs} refers to the first-person perspective inputs for the agent.

\subsection{Atari 100k \cite{Bellemare2012TheAL}}
\begin{figure*}[h]
\centering
\includegraphics[width=0.7\textwidth]{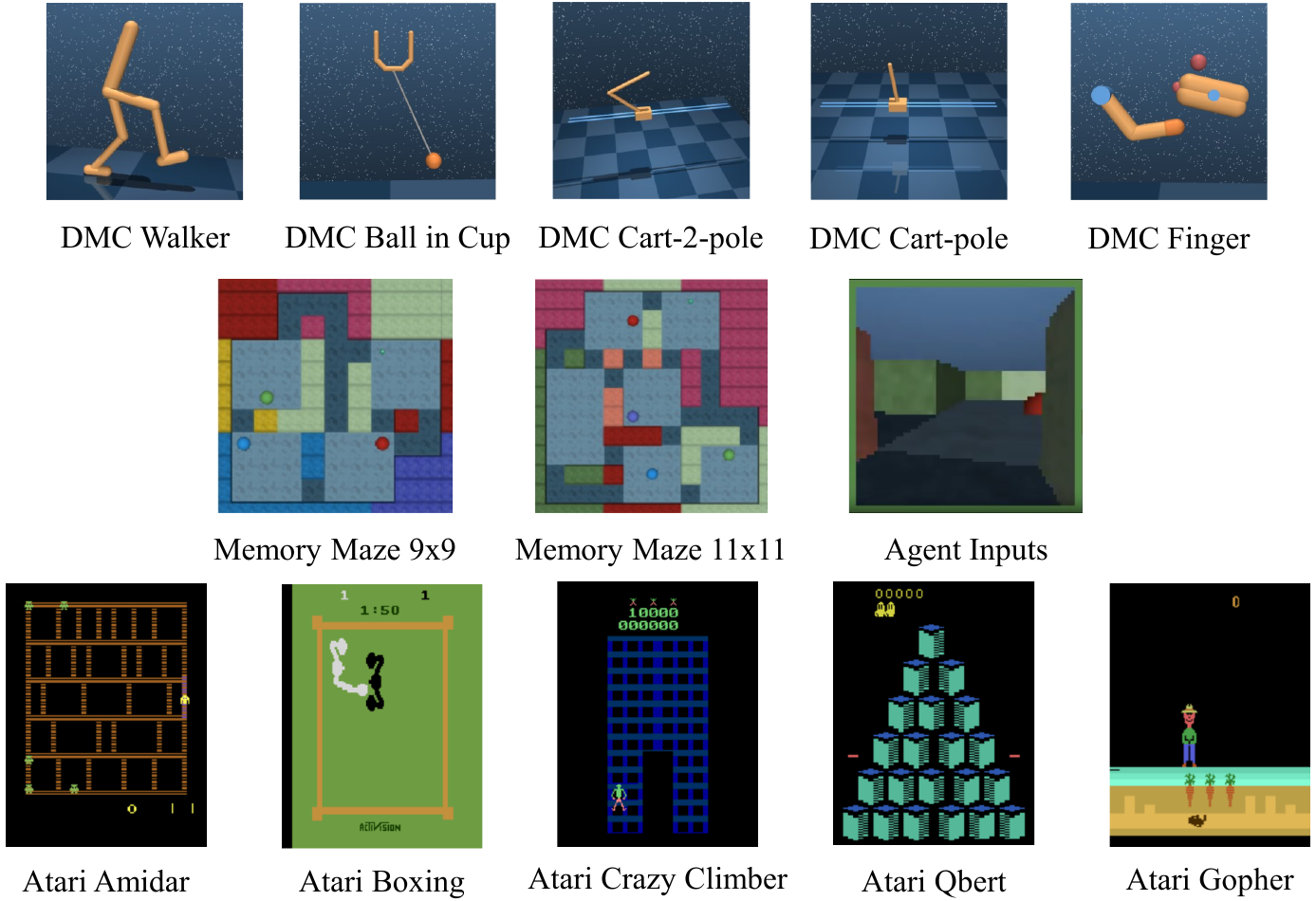} 
\caption{Tasks across three different domains are included in our paper to evaluate the effectiveness of APE.}
\label{fig:env}
\end{figure*}

The Atari 100k task contains 26 video games with up to 18 discrete actions, which are often serve as benchmarks for sample efficiency. Considering frame skipping (4 frames skipped) and repeated actions within those frames, the 100k sample constraint equates to 400k actual game frames. Given the wide domain gap between real-world images and Atari observations (reported in Appendix B), we consider five tasks in our evaluation. Visualized observations are illustrated in the third line of Fig. \ref{fig:env}.

\section{B\quad Additional Results}

\subsection{Comparison with PIE-G \cite{Yuan2022PreTrainedIE}}

We report the performance of APE against other ResNet pretrained algorithm in Table \ref{table:pieg}.
Results indicates that APE have better learning efficiency than other pretrained methods on both 100K and 500K environment step benchmarks and achieves comparable performance with SAC:state \cite{SAC} at 100K environment step. Results are averaged over at least 3 runs.



\begin{table*}[h!]
  \begin{center}

    \begin{tabular}{l|c|cc|ccc|cc} 
      \toprule  
      Task &\makecell[c]{SAC:\\state}& \makecell[c]{SAC:\\vision} & RRL & DrQ-v2& {PIE-G} & \makecell[c]{APE\\(DrQ-v2)} & DreamerV3 & \makecell[c]{APE\\(DreamerV3)}\\
      \midrule
      \midrule
      \multicolumn{3}{l}{\textit{100K Environment Step}} \\
      \midrule
      Walker Walk & \textbf{891} & 28 & 63 & 169.6 & 336.9 & 428.2 & 635.1 & \textbf{877.2}\\
      Finger Spin  & \textbf{811} & 158.8 & 135 & 325.2 & 539.9& 518.4 & 330 & 716.1\\
       Cup Catch & 746 & 177.5 & 261& 359& 587.9& 734& 410.8 & \textbf{916.8}\\
       \midrule
       \rowcolor{gray!20}
       Mean & \textbf{816}& 121.4& 153& 284.6& 488.2& 560.2& 458.6& \textbf{836.7} \\
       \midrule
        \multicolumn{3}{l}{\textit{500K Environment Step}} \\
        \midrule
        Walker Walk& \textbf{948} & 34.3 & 148& 704.7&689 & 680.5& \textbf{950.4} & \textbf{943.8}\\
        Finger Spin  & \textbf{923} & 296.8 & 422& 788.6 & \textbf{963.7} & 894.9& 439.2& 742.2\\
        Cup Catch & \textbf{974} & 639.4 & 447& 825.9 & \textbf{947.4}& \textbf{955.8}& 857.6 & \textbf{962.4}\\
        \midrule
        \rowcolor{gray!20}
        Mean & \textbf{948.3} & 323.5& 339& 773.1& 866.7 & 843.7& 749.1& 882.8\\
      \bottomrule 
    \end{tabular}
  \end{center}
  \caption{Comparison of APE against other ResNet pretrained algorithms (RRL \cite{RRL} and PIE-G) and their baselines (SAC:vision and DrQ-v2 \cite{Yarats2021MasteringVC}), together with SAC:state, which learns on proprioceptive observations.}
  \label{table:pieg} 
\end{table*}

\subsection{DMC Results}

Table \ref{tabledmc} shows the score of APE on DMC control tasks under 1M environment steps, 
compared with other state-of-the-art methods. The results of SAC, CURL, DrQ-v2, and DreamerV3
are from the paper of DreamerV3 \cite{dreamerv3} except for those used for visualization, whose "best" scores are reported, representing the best performance during training.
\begin{table*}[h!]
\centering

\begin{tabular}{l|ccccc}
\toprule
Tasks & SAC & CURL & DrQ-v2 & DreamerV3 & APE(Ours)\\
\midrule
\midrule
Cartpole Balance & \textbf{963.1} & \textbf{979} & \textbf{991.5} & $\textbf{999.8}$ & \textbf{998.8}\\
Cartpole Balance Sparse & \textbf{950.8} & \textbf{981} & \textbf{996.2} & \textbf{1000} & \textbf{1000}\\
Cartpole Swingup & 692.1 & 762.7 & \textbf{858.9} & 819.1 & \textbf{874}\\
Cartpole Swingup Sparse & \textbf{830.5} & 774.3 & 706.9 & $771.3$ & \textbf{845.2}\\
Cartpole Two Poles & 238& 255.4 & 295.8 & \textbf{437.6} & \textbf{482.8}\\
Cheetah Run & 27.2 & 474.3 & \textbf{691} & \textbf{728.7} & \textbf{688.6}\\
Cup Catch & 918.8 & \textbf{982.8} & 931.8 & $\textbf{981}$ & \textbf{985.5}\\
Finger Spin & 350.7 & 399.5 & 846.7 & $588.1$ & \textbf{969.9}\\
Finger Turn Easy & 176.7 & 338 & 448.4 & \textbf{787.7} & 721.6\\
Finger Turn Hard & 70.5 & 215.6 & 220 & \textbf{810.8} & \textbf{772.4}\\
Pendulum Swingup  & 560.1 & 376.4 & \textbf{839.7} &  \textbf{806.3} & \textbf{840.6}\\
Reacher Easy & 86.5 & 609.3 & \textbf{910.2} & 898.9 & \textbf{949.9}\\
Reacher Hard &  9.1 &  400.2 & \textbf{572.9} &  499.2 & 386\\
Walker Run & 26.9 & 376.2 & 517.1 & \textbf{757.8} & \textbf{758.2}\\
Walker Stand & 159.3 & 463.5 & \textbf{974.1} & \textbf{976.7} & \textbf{986.6}\\
Walker Walk & 268.9 & 909.4 & 762.9 & $\textbf{979}$ & \textbf{987.5}\\
\midrule
\rowcolor{gray!20}
Mean & 395.6 & 581.1 & 722.8 & \textbf{802.6} & \textbf{828}\\
\bottomrule
\end{tabular}
\caption{DMC scores for visual inputs after 1M environment steps.  }
\label{tabledmc}
\end{table*}

\subsection{Comparisons with ViT-Based Pretrained Encoder}
APE's efficacy lies in its augmentation strategy, outperforming methods merely rely on larger models or datasets. We finetuned MAE \cite{He2021MaskedAA}, a widely pretrained ViT encoder with diverse augmentations, to show APE’s effectiveness in three DMC tasks. Notably, APE achieved better results with much lower training time (19 GPU hours vs. 127.2 GPU hours for MAE). Results are shown in Table \ref{table:mae} (averaged over 3 runs).

\begin{table}[h!]
  \begin{center}
    
    \begin{tabular}{l|cc} 
      \toprule  
      Task & MAE (ViT)	 & APE (ResNet)\\
      \midrule
      \midrule
      DMC Mean 100K & 783.6 &\textbf{836.7} \\
      DMC Mean 500K & 809.1 & \textbf{882.8}\\

      \bottomrule 
    \end{tabular}
  \end{center}
  \caption{Comparisons with ViT-based pretrained encoder.}
    \label{table:mae} 
\end{table}

\subsection{Visualization of Reconstructions}
\begin{figure*}[h]
\centering
\includegraphics[width=0.65\textwidth]{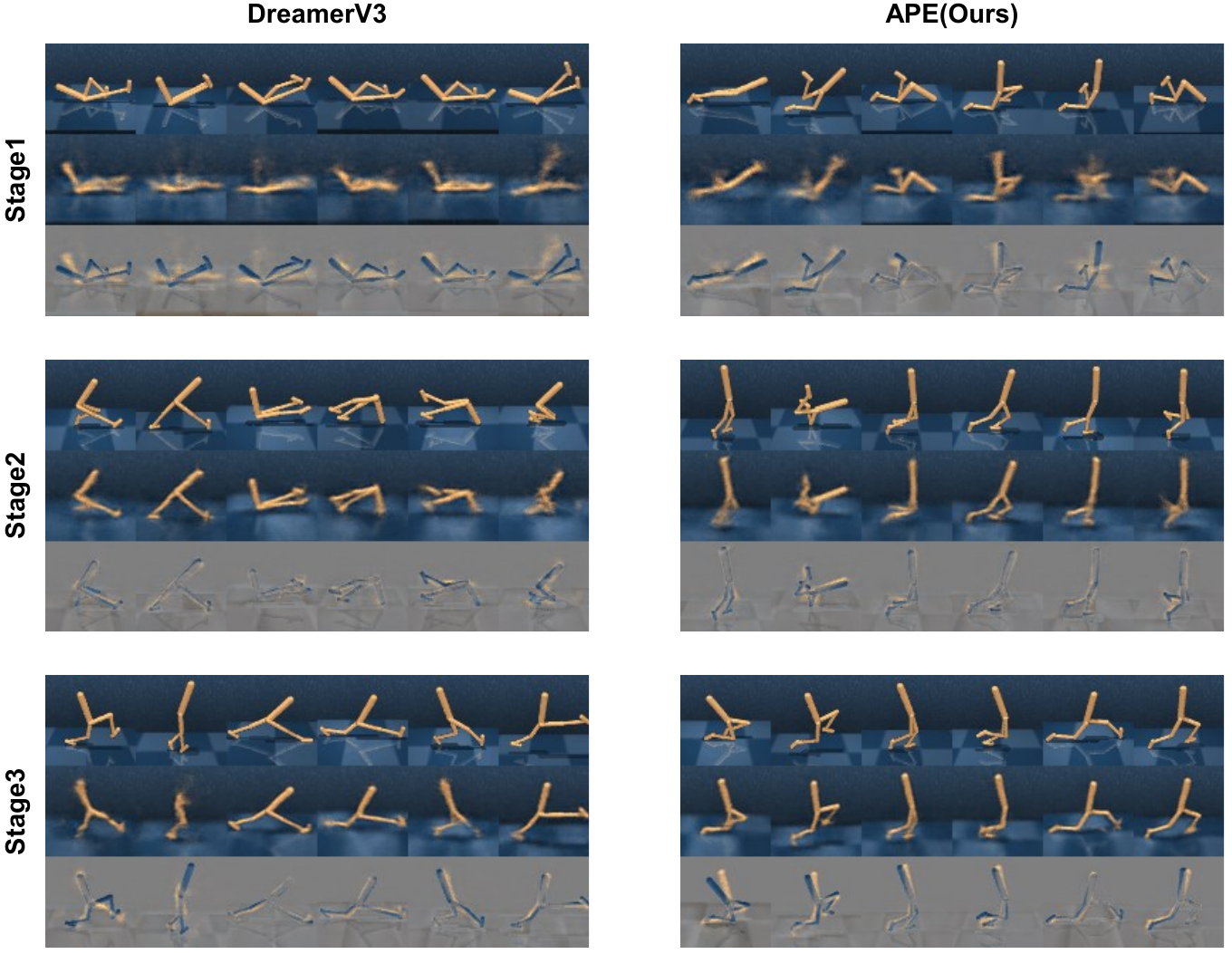} 
\caption{Visualization of reconstructions in different phases during policy learning period of \texttt{DMC walker walk}. The first row in each stage shows the real states of the agent, while the second row depicts the predictions reconstructed by the latent dynamics. The third row displays the prediction accuracy by comparing the actual states' outline with the predicted ones.}
\label{fig: recon}
\end{figure*}

As illustrated in Fig. \ref{fig: recon}, APE helps to perform more accurate predictions in the beginning of policy learning period (shown in Stage 1), enabling the agent to learn successful behaviors with fewer environment steps: APE manages to walk in Stage 2 while DreamerV3 struggles until Stage 3.

\subsection{Further Exploration on Atari Benchmarks}
\begin{wrapfigure}{R}{8cm}
\centering
\includegraphics[width=0.25\textwidth]{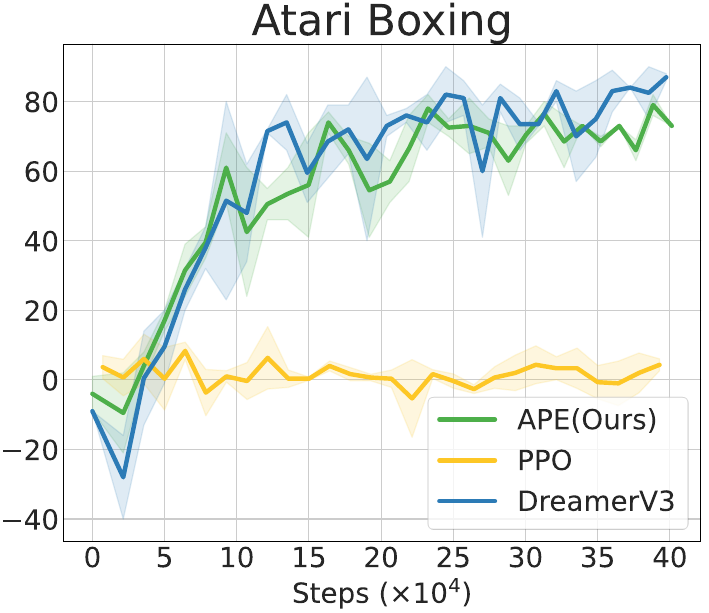}
\caption{Results on task with multi-item observations.}
\label{fig:img1}
\end{wrapfigure}
\begin{figure*}[h]
\centering
\includegraphics[width=0.8\textwidth]{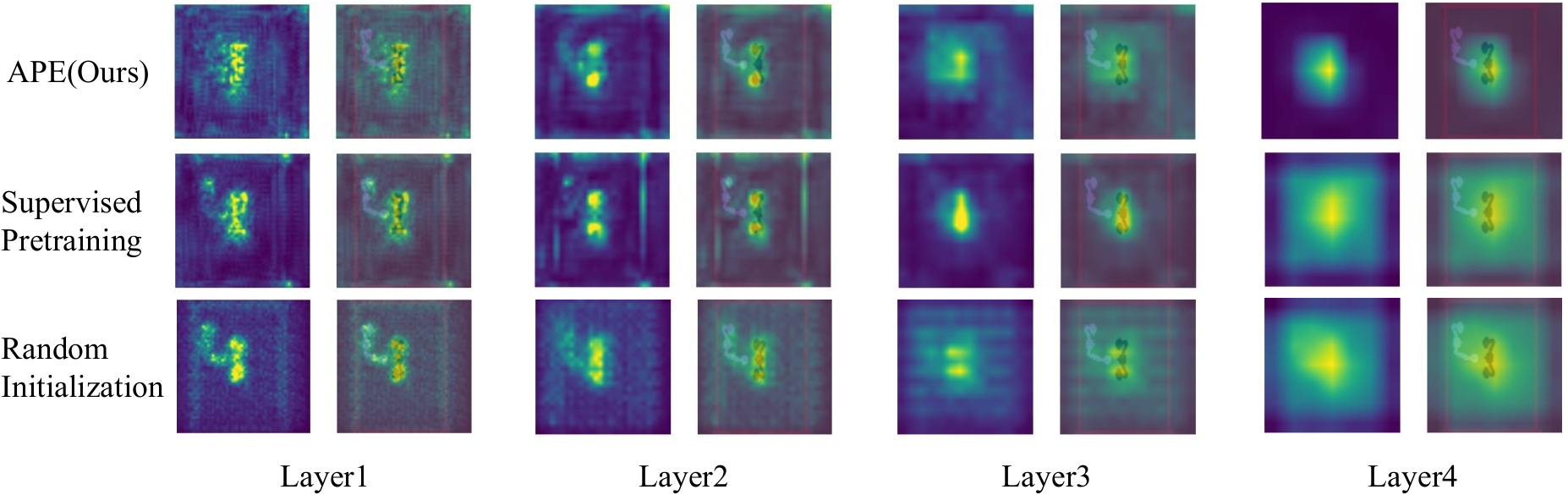} 
\caption{Visualization of different initialization of ResNet-18 model using LayerCAM \cite{9462463}.}
\label{fig: boxing}
\end{figure*}

We further explore the slight performance decrease of APE on several Atari tasks, e.g., \texttt{Atari Boxing} (shown in Fig. \ref{fig:img1}), where the agent is tasked to fight an opponent in a boxing ring. As illustrated in Fig. \ref{fig: boxing}, we visualize the features of different types of pretraining strategy to explore the generality of image classification models. 
For tasks with such challenging domain gap, APE achieves competitive results as supervised pretrained model, which is trained with a larger variety of images, i.e., ImageNet-1k. 
However, model with random initialization shows to be more adaptive to distributional shifts, since ImageNet-trained models are biased towards classifying single items instead of recognising multi-item observations, which are common in Atari tasks. In this case, the agent tends to overlook its opponents or targets, leading to a decline in Atari performance. We leave the improvement of multi-item detection in future work.

\section{C \quad Implementation Details}
\subsection{Agent Learning}
The actor and critic networks learn behaviors completely from the representations predicted by the latent dynamics, which produces a imagined sequence of states $s_t$, actions $a_t$, and continuation flags $c_t$. With $T$ represent the imagination horizon, the $\lambda$-return $G_t^\lambda$ \cite{Zhang2023STORMES} is computed as:
\begin{equation}\label{ec-9}
\begin{array}{l}
G_t^\lambda  \buildrel\textstyle.\over= {{r}_t} + \gamma {{ c}_t}\left[ {(1 - \lambda ){V_\psi }({s_{t + 1}}) + \lambda G_{t + 1}^\lambda } \right]\\
G_T^\lambda  \buildrel\textstyle.\over= {V_\psi }({s_T})
\end{array}
\end{equation}

We adopt the agent learning setting of DreamerV3 with the overall loss of the actor-critic algorithm remain unchanged, which can be described as follows \cite{Zhang2023STORMES}:
\begin{equation}
\begin{aligned}
    \mathcal{L}(\phi) = \frac{1}{T} \sum\limits_{t = 1}^T \left[{-\rm sg} \left(\frac{{ {G_T^\lambda-{V_\psi }({s_t})} }}{{\max (1,S)}} \right){\ln \pi_\phi({a_t}\mid{s_t}) } 
    - \eta H( {{\pi _\phi}({a_t}\mid{s_t})} ) \right] \\
    \mathcal{L}(\psi ) = \frac{1}{{T}}{ {\sum\limits_{t = 1}^T {\left[ \left({{V_\psi }({s_t}) - \rm sg\left( {G_t^\lambda } \right)} \right)^2  + \left({V_\psi }({s_t})-\rm sg\left(V_\psi^{EMA}(s_t)\right)\right)^2\right]} } }
\end{aligned}
\label{ec-8}
\end{equation}

\noindent where $\eta$ represent the coefficient for entropy loss, $H(\cdot)$ denotes the entropy of the policy distribution. The scale $S$ is used to normalize returns by:
\begin{equation}
    S = Per(G_T^\lambda, 95) - Per(G_T^\lambda, 5)
\end{equation}
\noindent here $Per(\cdot)$ computes an exponentially decaying average of the batch percentile.

Exponential moving average (EMA) is applied on updating the value function to prevent overfitting, which is defined as:
\begin{equation}
    \psi_{t+1}^{EMA} = \sigma\psi_t^{EMA} + (1-\sigma\psi_t)
\end{equation}
\noindent here $\sigma$ denotes the decay rate.
\subsection{Hyper Parameters and Setup for APE}
The pretext task trains for 200 epochs on 4 Nvidia Tesla A40 (48G) GPU servers while the evaluation runs for 100 epochs on 2 Nvidia Tesla A40 (48G) GPU servers. The RL agent is trained on one Nvidia Tesla A40 (48G) GPU server. Both the pretraining and policy learning algorithms are implemented using PyTorch’s packages. 

APE is pretrained on ImageNet-100, which is a subset of the common ImageNet-1k dataset \cite{Deng2009ImageNetAL}. It consists of 100 classes with a total of around 130,000 natural images, with each class containing roughly 1,000 images. This subset is often used for benchmarking and evaluating computer vision algorithms and models due to its diverse range of object categories and large number of images.

\begin{table}[h!]
  \begin{center}
    
    \begin{tabular}{l|cc} 
      \toprule  
      Environment & Action Repeat & Train Ratio\\
      \midrule
      \midrule
      DeepMind Control (DMC) & 2 &512 \\
      Memory Maze & 2 & 512\\
       Atari 100k & 4 & 1024\\

      \bottomrule 
    \end{tabular}
  \end{center}
  \caption{APE list of hyperparameters for each task.}
    \label{table:repeat} 
\end{table}

\begin{table}[h!]
  \begin{center}
    
    \begin{tabular}{l|c} 
      \toprule  
      {Hyperparameter} & Setting\\
      \midrule
      \midrule
     Input dimension & $3\times 224 \times 224$ \\
      Optimizer & SGD\\
       Learning rate & Res18\\
       \midrule
        \multicolumn{2}{l}{\textit{Pretext task}}\\
       \midrule
        Batch size & 128\\
        Learning rate & 3e-2\\
        Momentum & 0.999\\
        Weight decay & 1e-4\\
        Temperature & 0.2\\
        Queue & 65536\\
        \midrule
        \multicolumn{2}{l}{\textit{Linear Classification}}\\
       \midrule
       Batch size & 256\\
       Learning rate & 30\\
       Weight decay & 0\\
        \midrule
        \multicolumn{2}{l}{\textit{Data Augmentation}}\\
       \midrule
        $f_{Jitter}$ & 0.6, 0.7, 0.8 (default: 0.8)\\
        $f_{Blur}$ & 0, 0.2, 0.4, 0.5, 0.6, 0.8, 1 (default: 0.5)\\
        $f_{Flip}$ & 0.5 (default: 0.5)\\
        $f_{gray}$ & 0.2 (default: 0.2)\\
        Brightness delta& 0.4\\
        Contrast delta& 0.4 \\
        saturation delta& 0.4\\
        Hue delta & 0.1\\
      \bottomrule 
    \end{tabular}
  \end{center}
  \caption{APE list of hyperparameters in pretraining period.}
    \label{table:pre} 
\end{table}

\begin{table}[h!]
  \begin{center}

    \begin{tabular}{l|c} 
      \toprule  
      {Hyperparameter} & Setting\\
      \midrule
      \midrule
      Replay capacity & 1e6\\
     Input dimension & $3\times 64 \times 64$ \\
      Optimizer & Adam\\
      Batch size & 16\\
      Batch length & 64\\
      Policy and reward MPL number of layers & 2\\
      Policy and reward MPL number of units & 512\\
      Strides of the fourth layer for Res18 & 1, 1, 1, 1\\
      Strides of the fourth layer for Res50 & 1, 1, 1, 2\\
       \midrule
        \multicolumn{2}{l}{\textit{World Model}}\\
       \midrule
        RSSM number of units & 512\\
        Learning rate & 1e-4\\
        Adam epsilon & 1e-8\\
        Gradient clipping & 1000\\
        \midrule
        \multicolumn{2}{l}{\textit{Actor Critic}}\\
       \midrule
       Imagination horizon & 15\\
       Learning rate & 3e-5\\
       Adam epsilon & 1e-5\\
       Gradient clipping & 100\\
      \bottomrule 
    \end{tabular}
  \end{center}
  \caption{APE list of hyperparameters in policy learning period.}
    \label{table:policy} 
\end{table}

\begin{algorithm}[tb]
\caption{APE’s main training algorithm}
\label{alg:algorithm}
\textcolor{gray}{\texttt{//Adaptive Pretraining period}} \\
Initialize sampling probabilities ${\{p_i\}}_{i=1}^N$:

\qquad \qquad \qquad${p_1} = {p_2} = ... = {p_N}$

\begin{algorithmic}[1]

\FORALL {training epoch}
    \STATE compute the size of each sub-batch:\\
    $numbe{r_ - }dat{a_i} = soft\max (\alpha {p_i}) \times nu{m_ - }X$
    \STATE update samplers and resample sub-batches;
    \FORALL {sub-batches}
    \STATE draw two augmentation functions ${\Gamma _i}$ and ${{\Gamma '}_i}$;
    \STATE transform and map the training example;
    \STATE compute ${\mathcal{L}_z}$ and measure similarity;
    \STATE update networks to minimize ${\mathcal{L}_z}$;
    \STATE save the pretext task accuracy $ac{c_i}$;
    \ENDFOR
    \STATE update sampling probability for each sub-batch:\\
\qquad    $p_i^{t + 1} = Softmax(\alpha(1 - Acc_i^t))$
\ENDFOR 
\end{algorithmic}

\textcolor{gray}{\texttt{// Policy learning period}}\\
Initialize critic ${V_\psi }$ and actor ${\pi _w}$ and model ${M^\Delta }$

Loading pretrained encoder ${Encoder }$ with parameters $\varphi$
\begin{algorithmic}[1] 

\FORALL {$e=1,\cdots,E$}
            \STATE get initial state ${s_1} = {Encoder_\varphi }({o_1})$
            \FORALL {$t=1,\cdots,T$}
                \STATE obtain the latent feature ${s_t} = {Encoder_\varphi }({o_t})$
                \STATE apply action ${a_t} \sim {\pi _w}({a_t}|{s_t})$
                \STATE observe $s_{t+1}$ and $r_t$
                \STATE save transition $({s_t},{a_t},{s_{t + 1}},{r_t})$ in $\Re $
                \STATE generate $B$ random imaginary transitions of length $D$ starting from $s_t$ using ${M^\Delta }$
                \STATE store the imaginary transitions in $I$
                
                \FORALL {$k=1,\cdots,U_M$}
                    \STATE train ${M^\Delta }$ on minibatch from $\Re $
                \ENDFOR
                
                \FORALL {$k=1,\cdots,U_I$}
                    
                    \STATE train $\psi$ and $w$ on minibatch from $I $
                \ENDFOR       
            \ENDFOR
        \ENDFOR

\end{algorithmic}
\end{algorithm}
The DreamerV3-based APE is bulit upon the PyTorch DreamerV3 codebase\footnote{https://github.com/NM512/dreamerv3-torch} while the DrQ-v2-based APE is bulit upon the official PIE-G codebase\footnote{https://github.com/gemcollector/PIE-G/tree/master}.
Algorithm \ref{alg:algorithm} summarizes the training phase of APE. 
Hyperparameters for each task is provided in Table \ref{table:repeat}.
Moreover, we list the hyperparameters of the pretraining period and the policy learning period in Table \ref{table:pre} and Table \ref{table:policy} respectively.

\subsection{Hyper Parameters and Setup for Baselines}

Our PyTorch SAC implementation is based off of the official codebase\footnote{https://github.com/denisyarats/pytorch\_sac\_ae} of SAC+AE without decoder and thus achieves better performance than the common pixel SAC. The size of replay buffer for PIE-G is decreased to 50000 due to limited computational resources. The result of MoCo v2 with ResNet18 is bulit upon the official MoCo v2 codebase\footnote{https://github.com/facebookresearch/moco}.
\end{document}